\documentclass[11pt,a4paper]{article}
\usepackage{ifpdf}
\usepackage[bookmarks=false]{hyperref}
\usepackage[hyperref]{emnlp-ijcnlp-2019}
\usepackage{times}
\usepackage{latexsym}

\usepackage{url}

\usepackage{graphicx}
\usepackage{mathrsfs}
\usepackage{multirow}
\usepackage{amssymb}
\usepackage{amsmath}
\usepackage{booktabs}

\usepackage{array}
\newcolumntype{T}{!{\vrule width 1pt}}
\newlength\savedwidth
\newcommand\whline{\noalign{\global\savedwidth\arrayrulewidth
                            \global\arrayrulewidth 2pt}%
                   \hline
                   \noalign{\global\arrayrulewidth\savedwidth}}
\newlength\savewidth
\newcommand\shline{\noalign{\global\savewidth\arrayrulewidth
                            \global\arrayrulewidth 1pt}%
                   \hline
                   \noalign{\global\arrayrulewidth\savewidth}}

\aclfinalcopy 

\newcommand\confname{EMNLP-IJCNLP 2019}

\title{How to Evaluate the Next System: Automatic Dialogue Evaluation from the Perspective of Continual Learning}
\author{Lu Li\thanks{This work is done when the author is in Baidu.} \\
  Microsoft \\
  \And
  Zhongheng He\thanks{This work is done when the author is an intern in Baidu.} \\
  SHANNON.AI \\
  \And
  Xiangyang Zhou \\
  Baidu \\
  \And
  Dianhai Yu \\
  Baidu \\
  }
\date{2019-11-12}

\begin{document}
\maketitle

\begin{abstract}
Automatic dialogue evaluation plays a crucial role in open-domain dialogue research.
Previous works train neural networks with limited annotation for conducting automatic dialogue evaluation, which would naturally affect the evaluation fairness as dialogue systems close to the scope of training corpus would have more preference than the other ones.
In this paper, we study alleviating this problem from the perspective of continual learning: given an existing neural dialogue evaluator and the next system to be evaluated, we fine-tune the learned neural evaluator by selectively forgetting/updating its parameters, to jointly fit dialogue systems have been and will be evaluated.
Our motivation is to seek for a life-long and low-cost automatic evaluation for dialogue systems, rather than to reconstruct the evaluator over and over again.
Experimental results show that our continual evaluator achieves comparable performance with reconstructing new evaluators, while requires significantly lower resources.

\end{abstract}

\section{Introduction}

Automating dialogue evaluation is an important research topic for the development of open-domain dialogue systems.
Since existing unsupervised evaluation metrics, like BLEU \cite{papineni2002bleu} and ROUGE \cite{lin2004rouge}, are not suitable for evaluating open-domain dialogue systems \cite{liu2016not}, many researchers start to investigate conducting automatic dialogue evaluation by training-and-deploying: first, learn a neural network with annotation from existing dialogue systems, then use that trained evaluator to evaluate the other candidate dialogue systems \cite{kannan2017adversarial, lowe2017towards, tao2017ruber}. 
As most existing neural dialogue evaluators are only trained once, we call this kind of methods \emph{the stationary dialogue evaluators}.

\begin{figure}
\includegraphics[width = .43\textwidth,height=.23\textheight]{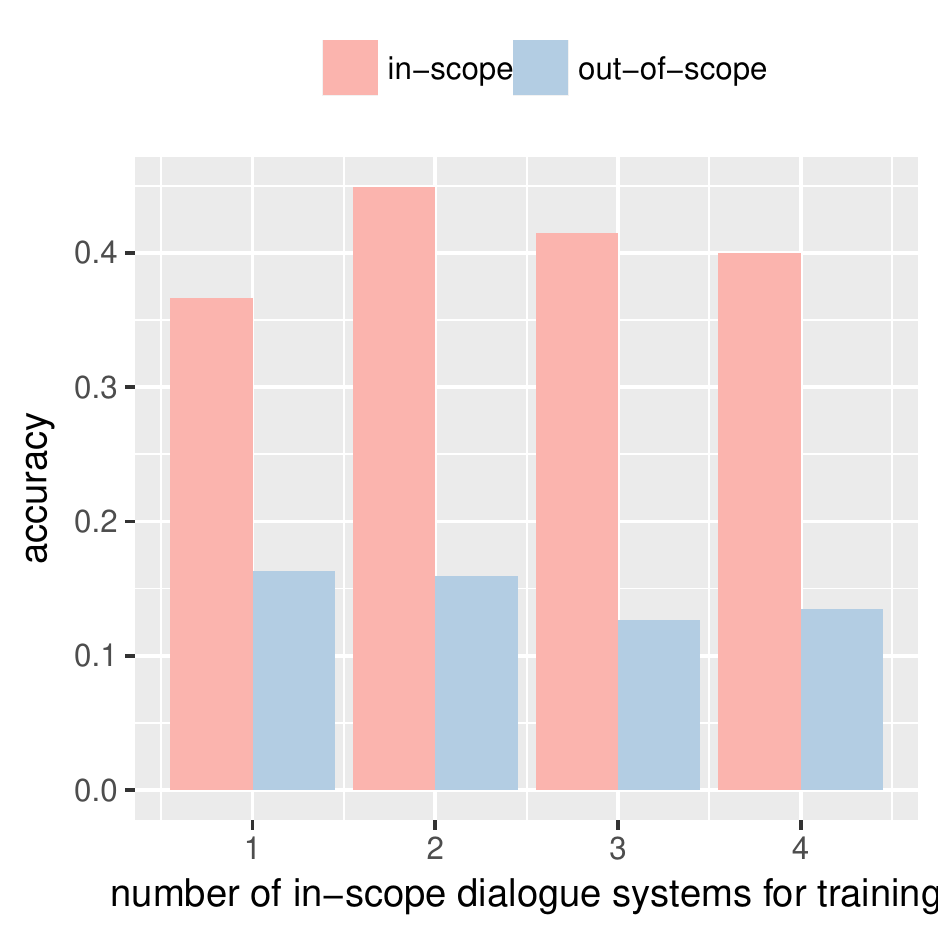}
\caption{\small
Performance gap of using different number of in-scope dialogue systems for evaluator learning. We got this result by using an existing neural dialogue evaluator, the ADEM \cite{lowe2017towards}, and prepared five different dialogue systems for comparison. The accuracy is the Spearman's rank correlation coefficient \cite{casella2002statistical} between human annotation and machine annotation, and the higher, the better.}

\label{fig-motivation}
\end{figure}

Though positive automatic evaluation results have been achieved by using stationary evaluators, we argue that the parameters of an evaluator learned from previous systems may not be able to fit future dialogue systems.
Figure \ref{fig-motivation} illustrates the performance gap of using the same stationary dialogue evaluator for two different groups of dialogue systems: 1) the \emph{in-scope systems} who are included in training corpus and 2) \emph{out-of-scope systems} that are excluded from evaluator learning.
As demonstrated, the automatic evaluation results of stationary neural dialogue evaluators are much more consistent to human annotation for systems that are involved in training than those outside the training scope.
Even  if increasing the number of dialogue systems for training the evaluator, the performance gap between in-scope and out-of-scope cannot be reduced.
Thus, responses from those out-of-scope systems could be more likely misestimated because the evaluator itself is not generalized enough to produce the correct prediction.
The flaws of generalization seriously limit the usability of neural dialogue evaluators, as the generalization is the key point for evaluation metrics. 


In this paper, we study fighting against the problem of generalization from the perspective of continual learning.
Let $e$ denote an existing neural evaluator, which is built for the previous $t$ dialogue systems, namely $\{d_1, d_2, ..., d_t\}$, given the next dialogue system to be evaluated, namely $d_{t+1}$, we extend the capacity of the evaluator $e$ by selectively fine-tuning its parameters with the next system $d_{t+1}$, in order to jointly fit dialogue systems in $\{d_1, ..., d_{t+1}\}$.
Unlike most fine-tuning methods that only care the fine-tuning side, we force the evaluator $e$ to learn from the next dialogue system $d_{t+1}$ without forgetting its knowledge learned from the previous systems.
By this way, the training scope of the evaluator is continually extended, including both dialogue systems in $\{d_1, ..., d_t\}$ and $d_{t+1}$, which can intuitively alleviate the weak generalization problem discussed above.

Particularly, we take the recently proposed automatic dialogue evaluation model (ADEM) \cite{lowe2017towards} as a base model, and explore fine-tuning it using two representative continual learning methods: the Elastic Weight Consolidation (EWC) \cite{kirkpatrick2017overcoming} and the Variational Continual Learning (VCL) \cite{nguyen2017variational}.
Both of those two continual learning algorithms can selectively update the model parameters, preserve those parts that are strongly related to the previous dialogue systems and rewrite the other parts in order to fit the next dialogue system.
Thus, the knowledge from the past and the future dialogue systems can coexist in the same evaluator, which enables the evaluator to treat all candidates equally.





The other solution to the weak generalization issue is reconstruction, either by simply re-building the evaluator using all annotations from $\{d_1, ..., d_{t+1}\}$ or by learning an additional one for $d_{t+1}$. However, this kind of methods naturally suffers from time, as we always need to maintain an increasing set of training annotation or an increasing set of evaluators whenever a new dialogue system comes, which could be too costly for long-term automatic evaluation.
However, by using continual learning, we only need to maintain one unified evaluator.
Besides, continually updating encourage knowledge transfer, which can reduce the size of annotation for the next dialogue system.
What is more, the continual learning based evaluator is more suitable for sharing, as people only need to open their evaluator instead of the raw training data, which can help protect the data privacy and safety.         

Experimentally, we build two sequences of dialogue systems, and each sequence consists of five different dialogue systems. 
We use different learning algorithms (including reconstructing, fine-tuning and continual learning) to sequentially update the base evaluator and evaluate each dialogue system one system after another.
Two major metrics are used to measure the performance of the automatic evaluation, i.e., 1) accuracy on evaluating the next dialogue system and 2) the consistency to its previous predictions, representing the plasticity and the stability respectively.
The comparison results show that the continual learning based evaluator is able to achieve comparable performance with other methods on evaluating the next dialogue system, while more stable and requiring significantly lower annotation.




Our major contributions are:
\begin{itemize}
\item We reveal the weakness of generalization in the previous neural dialogue evaluators. 
\item We propose solving the issue of generalization by incorporating two model-agnostic continual learning methods. Experimental results show that using continual learning can significantly alleviate the weakness of generalization.
\end{itemize}

\section{Problem Formalization and Methods}
\subsection{Problem Formalization}



Let $\{d_1, d_2, ..., d_T\}$ stand for a sequence of $T$ dialogue systems, a dialogue evaluator $e$ is asked to sequentially score those dialogue systems one after another.
At each step $t$ in the evaluation, a training set of post-reply-reference-label quads $\mathcal{D}_t = \{(c,r,g,l)_j^i\}_{i=1}^t$ collected from dialogue system $d_{t}$ would be available to update the evaluator for better automatic evaluation. The post $c$ is a single-turn conversational context, $r$ is the response generated by the corresponding dialogue system, $g$ is the reference response generated by a human, and label $l$ has three grades (2: good, 1: fair, 0: bad) and is manually annotated. 
After updating, the evaluator is asked to evaluate both the current dialogue system $d_t$ as well as the previous ones $\{d_1, ..., d_{t-1}\}$. 
The whole process of sequential evaluation can be formulated as:
\begin{align}
e_t = update(e_{t-1}, \mathcal{D}_{1:t})  \\
s_j^i(t) = score(c_j^i, r_j^i, g_j^i | \theta_t)
\end{align}
\noindent
where $\theta_t$ is the parameter of evaluator $e_t$ and 
$s_j^i(t) \in \mathbb{R}$ 
is the quality score generated by the evaluator $e_t$ for the $j^{th}$ post-reply-reference triple of the dialogue system $d_i$. The higher the score is, the better the response is. All comparison learning methods $update(\cdot)$ share the same evaluator architecture, but having their own parameters. In the following sub-sections, we will have a detailed description of the architecture of our evaluator $score(\cdot)$ as well as different learning algorithms $update(\cdot)$.

\subsection{Methods}

\subsubsection{Automatic Dialogue Evaluator}

\begin{figure}
\includegraphics[width = .48\textwidth,height = .11\textheight]{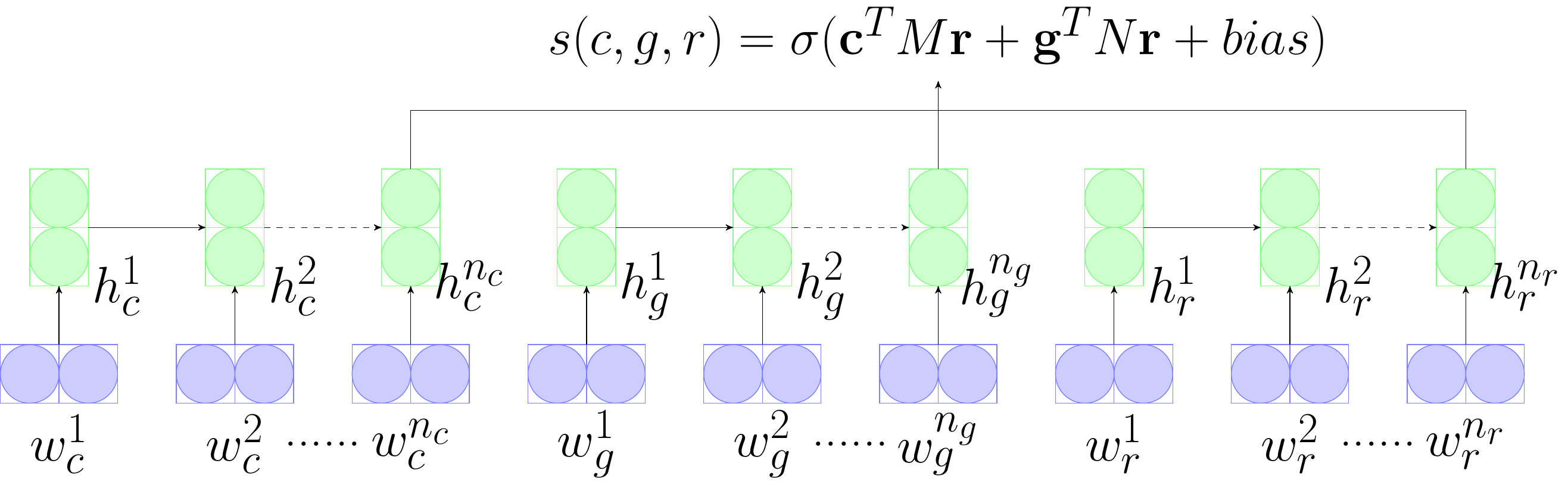}
\caption{\small The structure of ADEM.}
\label{fig-adem}
\end{figure}

Figure \ref{fig-adem} shows the structure of our shared neural dialogue evaluator, the ADEM. 
The idea of ADEM is to automatically measure the quality of the candidate response $r$ by jointly considering the dialogue context $c$ and the reference $g$. 
ADEM first leverages a shared 
LSTM cell
to read the context $c$, machine response $r$ and reference response $g$
 then uses the last hidden states 
 as the vector representations for $c$, $r$ and $g$, written as $\overrightarrow{c}$, $\overrightarrow{r}$ and $\overrightarrow{g}$ respectively.


Given that vector representations, the quality of the machine response $r$ is calculated as:
\begin{equation}
score(c, g, r | \theta) = \sigma(\overrightarrow{c}^TM\overrightarrow{r} + \overrightarrow{g}^TN\overrightarrow{r} + b)
\label{score-func}
\end{equation}
\noindent
where $M$ and $N$ are two learnable matrixes to measure the semantic similarities between the machine response $r$ and dialogue context $c$ as well as the reference response $g$, $b$ is the learnable bias, and $\theta$ denotes all parameters in this evaluator.

\begin{figure*}
\centering
\includegraphics[width = 0.9\textwidth, height= 0.25\textheight]{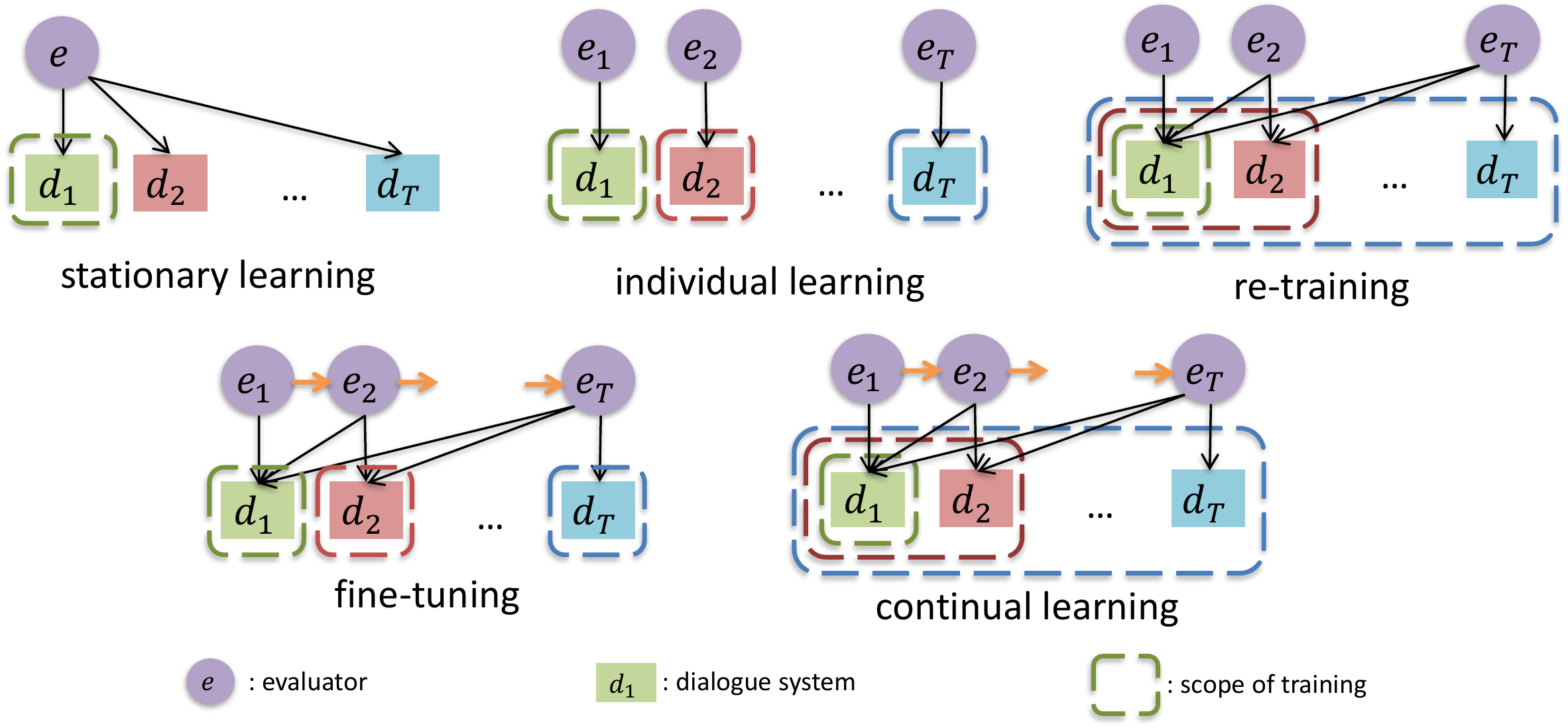}
\caption{\small Different learning processes of automatic dialogue evaluation.}
\label{fig-methods}
\end{figure*}

\subsubsection{Learning Methods}

Figure \ref{fig-methods} shows the framework of all investigated learning methods in this paper, including both the comparison baselines and our continual learning based methods.
Our baseline methods include four straight forward solutions:

\begin{description}
\item \textbf{stationary learning} only learns the evaluator once with the first dialogue system. 
\item \textbf{individual learning} trains multiple evaluators, each evaluator is tuned with and working for one specific dialogue system.
\item \textbf{retraining} re-learns the evaluator over and over again, gradually increase its training scope each time.
\item \textbf{fine-tuning} also only maintains the evaluator over time, however, at each time $t$ of updating, it only leverages the annotation from the current dialogue system $d_t$ and reuses the learned weights from $\{d_1, d_2, ..., d_{t-1}\}$. As fine-tuning only cares the current dialogue system $d_t$, its training scope only shifts across different dialogue systems, rather than increasing.
\end{description}

Unlike baseline methods, the continual learning algorithm can gradually increase the training scope of neural networks without retraining from scratch. 
Particularly, at each time $t$ of updating, the continual learning algorithm only requires the annotation of the current dialogue system $d_t$, similar to fine-tuning.
However, instead of overwriting all parameters like fine-tuning, the parameters of the neural dialogue evaluator are selectively modified, those weights that are strongly related to the previous dialogue systems are consolidated while the other parts are overwritten to fit the new dialogue system. 
Hence, the training scope of continual learning can be gradually increased after multiple steps of updating.
We apply two kinds of continual learning methods for automatic dialogue evaluation: the Elastic Weight Consolidation (EWC) \cite{kirkpatrick2017overcoming} and the Variational Continual Learning (VCL) \cite{nguyen2017variational}.



Given a dialogue evaluator $e_t$ and its parameters $\theta_t$, \textbf{EWC} assumes that the importance of weights $\theta_t$ are not equal, some of those parameters are strongly contributing to the prediction while the others are not.
The difference of parameter importance enables the model to learn new knowledge without forgetting the old ones, as we can force those parameters, that has less contribution to previous systems, to store more knowledge for the new dialogue system. 
To measure the contribution of parameters, EWC exploits the Fisher Information Matrix \cite{frieden2004science}, which is calculated as:
\begin{align}
F_i = E[ (\frac{\delta L}{\delta \theta_i})^2 | \theta]
\end{align}
\noindent
where $L$ denotes some certain loss function and
 $\theta_i$
 is the $i^{th}$ weight in the parameter $\theta$. The Fisher Information $F_i$ measures the importance of $\theta_i$ to the object $L$. 

Given the Fisher Information Matrix $F^{t-1}$, EWC updates the the evaluator with the following object $L_{EWC}(\theta_t)$ at each time $t$:

{\small
\begin{align}
L_{EWC}(\theta_t) = L( \mathcal{D}_t, \theta_{t}) + L(\theta_{t}, \theta_{t-1}^*) \\
L( \mathcal{D}_t, \theta_{t}) = \frac{1}{J} \sum_j (l_j^t - score(c_j^t, r_j^t, g_j^t | \theta_t))^2 \\
L(\theta_{t}, \theta_{t-1}^*) =  \sum_i \frac{\lambda}{2} F_i^{t-1}(\theta_{t, i}-\theta_{t-1, i}^*)^2 \\
F_i^{t-1} = E\{ (\frac{\delta L( \mathcal{D}_{t-1}, \theta_{t-1})}{\delta \theta_{t-1, i}})^2 \}
\label{ewc-all}
\end{align}
}

\noindent
where  
$\theta_{t-1}^*$ is learned value of $\theta_{t-1}$ at time $t-1$, 
$L( \mathcal{D}_t, \theta_{t})$ is the regression object for fitting the data of the dialogue system $d_t$ and 
$L(\theta_{t}, \theta_{t-1}^*)$ is the object to memorize the important weights for the previous systems. 
$F_i^{t-1}$ is the importance score for the $i^{th}$ weight in the parameter for previous $t-1$ dialogue systems, and 
$(\theta_{t, i}-\theta_{t-1, i}^*)^2$ 
measures the variance of the $i^{th}$ parameter. The $score(c_j^t, r_j^t, g_j^t | \theta_t)$ represents the prediction of evaluator $e_t$ for the $j^{th}$ post-response-reference triplet of the dialogue system $d_t$, $l_j^t \in [0, 1, 2]$ is the label of that instance. 
The definition of $score(c_j^t, r_j^t, g_j^t | \theta_t)$ is in Equation ~\ref{score-func}. $\lambda$ is a hyper-parameter for trading off those two losses.
With that joint loss, the EWC algorithm can force the neural dialogue to learn without forgetting, gradually increasing its training scope step by step.

Different to EWC, VCL \cite{nguyen2017variational} assumes that all parameters in the model are random variables, 
whose prior distribution is $p(\theta)$.
Given a sequence of $T$ dialogue systems $\{d_1, d_2, ..., d_T\}$, the VCL based dialogue evaluator learns the weights for all the $T$ systems by Bayesian rule:

\begin{equation}
\begin{aligned}
p(\theta|d_{1:T}) &= \frac{p(\theta) \cdot p(d_{1:T}|\theta)}{p(d_{1:T})} \\
&\propto p(\theta) \cdot p(d_{1:T-1}\cup d_T|\theta) \\
&\propto p(\theta)\cdot\frac{p(\theta|d_{1:T-1})p(d_{1:T-1})}{p(\theta)}\cdot p(d_T|\theta)\\
&\propto p(\theta|d_{1:T-1}) \cdot p(d_T|\theta)
\label{recursion}
\end{aligned}
\end{equation}
\noindent
where, $p(\theta|d_{1:T})$ denotes the distribution of parameters learned from $\{d_1, ..., d_T\}$, and $p(d_t|\theta)$ denotes the generation probability of using $\theta$ to fit the dialogue system $d_t$. 
The above formula implies that we can naturally learn the joint neural evaluator for all dialogue systems in $\{d_1, ..., d_T\}$ via continually transferring knowledge from previous dialogue systems, i.e., first learning $p(\theta | d_1)$, then learning $p(\theta | d_1, d_2)$, and finally learning $p(\theta|d_{1:T})$.


However the posterior distribution $p(\theta | d_{1:T})$  is intractable, hence we use $q_t(\theta)$ to approximate it  through a KL divergence minimization \cite{sato2001online}, 
\begin{equation}
\small
q_t(\theta) = arg min_{q \in \mathcal{Q}} KL(q(\theta) || \frac{1}{Z_t}q_{t-1}(\theta)p(D_t|\theta))
\label{KL}
\end{equation}
for $t \ge 1$.
The initial distribution $q_0(\theta)$ at time 0 is defined as the prior distribution $p(\theta)$, and $Z_t$ is the normalizing constant.
Follow the existing work \cite{friston2007variational}, minimizing the KL divergence (eq.\ref{KL}) is equivalent to maximizing the negative variational free energy so that the loss function can be defined as:
\begin{equation}
\small
L_{VCL}^t(q_t(\theta)) = E_{\theta \sim q_t(\theta)} [L(D_t, \theta)] + KL(q_t(\theta)||q_{t-1}(\theta))
\end{equation}
where $L( \mathcal{D}_t, \theta)$ is the regression loss define in Equation ~\ref{ewc-all}.
Thus we can learn $q_t(\theta)$ sequentially from $q_1(\theta)$, $q_2(\theta)$ and $q_{t-1}(\theta)$.
During prediction at each time $t$, we first sample a set of parameters from $q_t(\theta)$ and then use
the mean prediction to evaluate the quality of dialogue systems.

\section{Experiment}

\subsection{Data}
To test the performance of using continual learning for automatic dialogue evaluation, we build five different dialogue systems for simulating automatic evaluation, including a typical retrieval-based system (retrieval-system \cite{yan2016docchat}), three generation-based systems (vanilla seq2seq with attention model \cite{bahdanau2014neural}, seq2seq with keyword model \cite{yao2017towards}, CVAE-based seq2seq model \cite{yao2017towards}) and a user simulator, which responds by crowd-sourcing. 
The raw dataset consists of multiple quads $<$single-turn post, machine response, reference response, label$>$.
The posts and reference responses are collect from Baidu Tieba\footnote{http://tieba.baidu.com}, which is the largest Chinese online forum in open-domain topics.
Given that post set, our five comparison systems are asked to produce responses.
After responses have been generated, five annotators are asked to rate each candidate response according to the given single-turn post and the reference response.
The quality score has three grades: 0 (bad), 1 (fair) and 2 (good), and we choose the score of each response via voting.
All five systems are anonymous so that annotators do not know which system the response is from during annotation.
After annotation, we totally collect 30k $<$single-turn post, machine response, reference response, label$>$ quads, we further randomly split that data into training set(20k), validation set (5k) and test set (5k) by post, i.e., different responses replying to the same pose only belongs to one set.

\subsection{Metrics}
We measure the performance of automatic evaluation from two different perspectives: 1) \emph{generalization} to the next dialogue system and 2) \emph{consistency} to the previous predictions.
An ideal evaluator with good generalization and consistency shall be able to provide a life-long automatic evaluation.
Same as most previous works \cite{liu2016not, lowe2017towards, tao2017ruber}, we utilize Spearman's rank correlation coefficient \cite{casella2002statistical} to calculate generalization and consistency.

Given the predictions $\{s^i_j(t)\}_{j=1}^{J}$ of evaluator $e_t$ for the $J$ instances of dialogue system $d_i$, we calculate its similarity to the human annotation $\{l^i_j\}_{j=1}^{J}$ as:

\begin{equation}
\small
sp(e_t, h, d_i) = \frac{\sum_{j=1}^{J} (s_j^i(t) - \overline{s^i}(t))(l_j^r - \overline{l^i})}{
\sqrt{\sum_{j=1}^{J}(s_j^i(t) - \overline{s^i}(t))^2}
\sqrt{\sum_{j=1}^{J}(l_j^i - \overline{l^i})^2}
 }
\end{equation}
\noindent
where $\overline{s^i}(t)$ and $\overline{l^i}$ are the mean value of $\{s^i_j(t)\}_{j=1}^{J}$ and $\{l^i_j\}_{j=1}^{J}$.

Spearman's rank correlation can provide a convenient way for us to calculate generalization and consistency, specifically, we define generalization and consistency as:

\begin{align}
Pla^t = sp(e_t, h, d_t) \\
Sta^t = \frac{1}{t-1}\sum_{i=1}^{t-1} sp(e_i, e_t, d_i)
\end{align}
\noindent
where $Pla^t$ computes the correlation with human annotation, for the evaluator $e_t$'s prediction on the next dialogue system $d_t$, standing for the generalization (plasticity).
While $Sta^t$ is the average of the correlation of the prediction at time $t$ with its previous prediction in range of $[1, t-1]$, standing for the consistency (stability).

\begin{table*}
	\small
	\centering
	\begin{tabular}{cTcTc|c|c|c|cTc|c|c|c}
	        \whline
		\multirow{2}*{order} & \multirow{2}*{methods} & \multicolumn{5}{cT}{Plasticity} & \multicolumn{4}{c}{Stability}  \\
		\cline{3-11}
		~&~& $t=1$ &$ t=2$ & $t=3$ & $t=4$ & $t=5$  & $t=2$ & $t=3$ & $t=4$ & $t=5$\\
		\shline
		\multirow{6}*{in time}&stationary learning & 0.370 & 0.247 & 0.236 & 0.176 & 0.129 & 1.000 & 1.000 & 1.000 & 1.000 \\
		\cline{2-11}
		~&individual learning & 0.370 & 0.488 & 0.459 & 0.337 & 0.164 & 0.516 & 0.680 & 0.642 & 0.488 \\
		\cline{2-11}
		~&retraining & 0.366 & 0.451 & 0.398 & 0.298 & 0.282 &0.751 & 0.743 & 0.691 & 0.657 \\
		\cline{2-11}
		~&fine-tuning & 0.376 & 0.485& 0.449 & 0.345  & 0.380& 0.549 & 0.719 & 0.744 & 0.506  \\
		\cline{2-11}
		~&EWC & 0.367 & 0.488 & 0.458 & 0.338 & 0.346 & 0.587 & 0.767 & 0.761 & 0.688  \\
		\cline{2-11}	
		~&VCL &0.362 &0.365 & 0.333& 0.230 &0.331& 0.934 & 0.930 & 0.870 & 0. 741  \\
                \shline
	        \multirow{6}*{in random}&stationary learning& 0.370 & 0.248 & 0.363 & 0.157 & 0.139 & 1.000 & 1.000 & 1.000 & 1.000  \\
	        \cline{2-11}
		~&individual learning & 0.164 & 0.459 & 0.488 & 0.337 & 0.370 & 0.449 & 0.667 & 0.604 & 0.414  \\
		\cline{2-11}
		~&retrain & 0.370 & 0.427 & 0.458 & 0.295 & 0.262& 0.803 & 0.779 & 0.674 & 0.658\\
		\cline{2-11}
		~&fine-tuning &  0.370&0.442 & 0.488& 0.349 & 0.390& 0.611 & 0.754 & 0.657 & 0.519  \\
		\cline{2-11}
		~&EWC & 0.379 & 0.439 & 0.492 & 0.326  & 0.350& 0.692 & 0.801 & 0.681 & 0.688  \\
		\cline{2-11}	
		~&VCL &0.368 &0.322 & 0.434 & 0.224  & 0.238  & 0.939 & 0.945 & 0.819 & 0.856\\
	        \whline
	\end{tabular}
	\caption{\small Experimental results for evaluating dialogue systems.}
	\label{Results} 
\end{table*}

\subsection{Training}
In order to make the learning methods more comparable, we make sure that all hyper-parameters and data-preprocess are same for all comparison methods.
For the variable-size input, we use zero pads for shorter input and cut the longer input if its length exceeds 50. 
The vocabulary size is about 0.48 million cut by word frequency of the training set for pre-training.
All models are trained by AdamOptimizer \cite{kingma2014adam} with learning rate 0.001 and batch size 32.

For the base model, the embedding size is 256 and the RNN cell is one layer LSTM cell with hidden size 512.
Similar to the previous work \cite{lowe2017towards}, we use a pre-training procedure to learn the parameters of the encoder. In this work, we train the encoder as part of a matching model. The last layer parameters are initialized from a truncated normal distribution with a mean of 0 and a standard deviation of 0.1.

For stationary learning, we use the training/validation set of the first dialogue system to train and choose models. 
Once learned stationary dialogue evaluator, we fix it to evaluate all incoming dialogue systems.
For individual learning multiple evaluators, we train/choose individual evaluator only according to the training/validation set of the individual dialogue system.
For retraining, we use accumulated validation set of all seen tasks to choose the best model due to the training set is also accumulated when every new dialogue system emerges.
For fine-tuning, we update/choose model using training/validation set of new dialogue system while init the model parameters by previous evaluator parameters.
For continual learning, the training and model choosing processes are same with fine-tuning but add some penalty during training.
The lambdas of EWC are $[10^4, 10^5, 10^6, 10^7]$, increasing the penalty by ten times with the number of task increases. 
In VCL, the prior distribution of model variables is Gaussian distribution with a mean of 0 and a variance of $1e^{-6}$ same as precious work \cite{nguyen2017variational}, and the sample number is 20. 

In order to ensure the robustness of the experiment, we simulate two sequences of dialogue systems with five different dialogue systems in two orders: 1) time order and 2) random order.
A sequence of dialogue systems in time order, i.e. the retrieval-system $\to$ the seq2seq with attention model $\to$ the seq2seq with keywords model  $\to$ CVAE-based seq2seq model $\to$ human simulator, imitates the reality that new dialogue systems are constantly emerging with the development of dialogue technology. 
Additionally, we randomly sort the five dialogue systems as the human simulator $\to$ seq2seq with keywords model $\to$ seq2seq with attention model $\to$ CVAE-based seq2seq model $\to$ the retrieval system, as a sequence of dialogue systems in random order.
We evaluate the two dialogue system sequences using baseline methods and continual learning methods.
\subsection{Results}
Table \ref{Results} shows the results in evaluating two dialogue system sequences, as demonstrated, the two sets of experiments yield consistent results.
On the plasticity (generalization) side, fine-tuning and EWC achieve the best performance while stationary learning is the worst, proving our motivation on the generalization issue of existing stationary dialogue evaluators.
Retraining is lower than fine-tuning, which we believe is because retraining cares all systems but fine-tuning only concerns the current ones.
It is interesting to find that individual learning is not as good as fine-tuning, which we think the reason is that fine-tuning can benefit from the knowledge of previous systems, so do EWC and VCL.

On the stability (consistency) side, stationary learning is the best one as it never changes. 
Among all adaptive methods, retraining and VCL achieve the best stability scores, followed by EWC, fine-tuning and individual learning are the worst as they only care their own systems. 

Continual learning based evaluators (EWC and VCL) outperform other baselines, jointly considering generalization and consistency.
EWC is more plastic than VCL, while VCL is much more stable.
We believe this is because VCL is a probabilistic neural network, which naturally considers varieties in its model and thus has stronger stability, which however increases the difficulty in training.

\section{Analysis}
\subsection{Complexity}
We analyze time and space complexity during the whole training-and-deploying process, see Table \ref{complexity}.
We ignore the complexity of model architecture and the difference in learning methods. We assume that updating one evaluator through one training set takes exactly one time step. 
Restore the labeled data of one dialogue system and one evaluator cost one space.

For evaluating dialogue system sequence consisting of $n$ systems,
training stationary dialogue evaluator only cost $O(1)$ time to evaluate all systems and there is only one evaluator and no data need to restore.
Training one individual evaluator and sequential updating (fine-tuning/continual learning) evaluator take $O(1)$ time when very next dialogue system emerging, thus total cost O(n) time.
For retraining, when every new dialogue evaluator emerging, we need to train all previous system, thus would cost total $O(n^2)$ time and need to restore all data for all existing systems.  
For source restore, stationary learning and sequential updating only need to retrain one latest evaluator, while individual learning need to restore all learned evaluators and retraining need to save all training data.
Despite stationary learning, sequential updating (fine-tuning/continual learning) is a better method, in terms of time and space complexity.

\begin{table} 
	\small
	\begin{tabular}{c|c|c}
	\hline
	methods & time complexity & space complexity \\
	\hline
	stationary learning & $O(1)$ & $O(1)$ \\
	\hline
	individual learning & $O(n)$ & $O(n)$ \\
	\hline
	sequential updating & $O(n)$ & $O(1)$ \\	
	\hline
	retraining & $O(n^2)$ & $O(n)$ \\
	\hline
	\end{tabular}
	\caption{\small Time and space complexity.}
	\label{complexity} 
\end{table}

\subsection{Stability-plasticity dilemma}

\begin{table*}
        \scriptsize
	\centering
	\begin{tabular}{cTc|c|c|c|cTc|c|c|c|cTc|c|c|c|c}
		\shline
		\multirow{2}*{}& \multicolumn{5}{cT}{fine-tuning} &  \multicolumn{5}{cT}{EWC} & \multicolumn{5}{c}{VCL} \\
		~& \multicolumn{5}{cT}{$r \to a \to k \to c \to h $} &  \multicolumn{5}{cT}{$r \to a \to k \to c \to h $} & \multicolumn{5}{c}{$r \to a \to k \to c \to h $}\\
		\shline
		$h$ &  &  &  &  & 0.380 &        &  &  &  & 0.346 &                     &  &  &  & 0.331\\
		\hline
		$c$ &  &  &  & 0.345 & 0.190 &         &  &  & 0.338 & 0.260 &          &  &  & 0.230 & 0.196\\
		\hline
		$k$ &  &  & 0.449 & 0.416 & 0.305 &        &  & 0.458 & 0.426 & 0.409 &            &  & 0.333 & 0.311 & 0.291\\
		\hline
		$a$ &  & 0.485 & 0.454 & 0.467 & 0.415 &        & 0.488 & 0.483 & 0.472 & 0.460 &            & 0.365 & 0.394 & 0.371 & 0.388\\
		\hline
		$r$ & 0.376 & 0.245 & 0.314 & 0.267 & 0.168 &            0.367 & 0.255 & 0.313 & 0.252 & 0.205 &        0.362 & 0.349 & 0.340 & 0.320 & 0.232\\
		\shline
	\end{tabular}
	\caption{\small The plasticity for evaluating dialogue systems using fine-tuning, EWC, VCL in time order. The lowercase letters are short for dialogue systems, and $r$, $a$, $k$, $c$, $h$ indicate retrieval-system, attention-system, keywords-system, CVAE-system, human-system respectively.}
	\label{dilemma}
\end{table*}

Table \ref{dilemma} shows the plasticity on different dialogue systems when using fine-tuning and continual learning.
Different from fine-tuning, continual learning (EWC and VCL) add restrictions on the learning of the next task, which would reduce the interference on the previous task but affect the learning capacity of the new task.
For example, after learning from human-system, the accuracy on human-system of fine-tuning based dialogue evaluator is the highest 0.380, while the accuracy on CVAE-system decreases sharply from 0.345 to 0.190.
Although for EWC and VCL based dialogue evaluators, the accuracy on CVAE-system is only slightly affected, but the plasticity of human-system decrease.
 This is the stability-plasticity dilemma.
The EWC based evaluator improves the stability meanwhile maintaining the plasticity.
To some extent, we consider that EWC based dialogue evaluator achieves a trade-off between stability and plasticity and alleviates the stability-plasticity dilemma.

\subsection{Supervision requirement}
\begin{figure}
\centering
\includegraphics[width = 0.45\textwidth, height = 0.26\textheight]{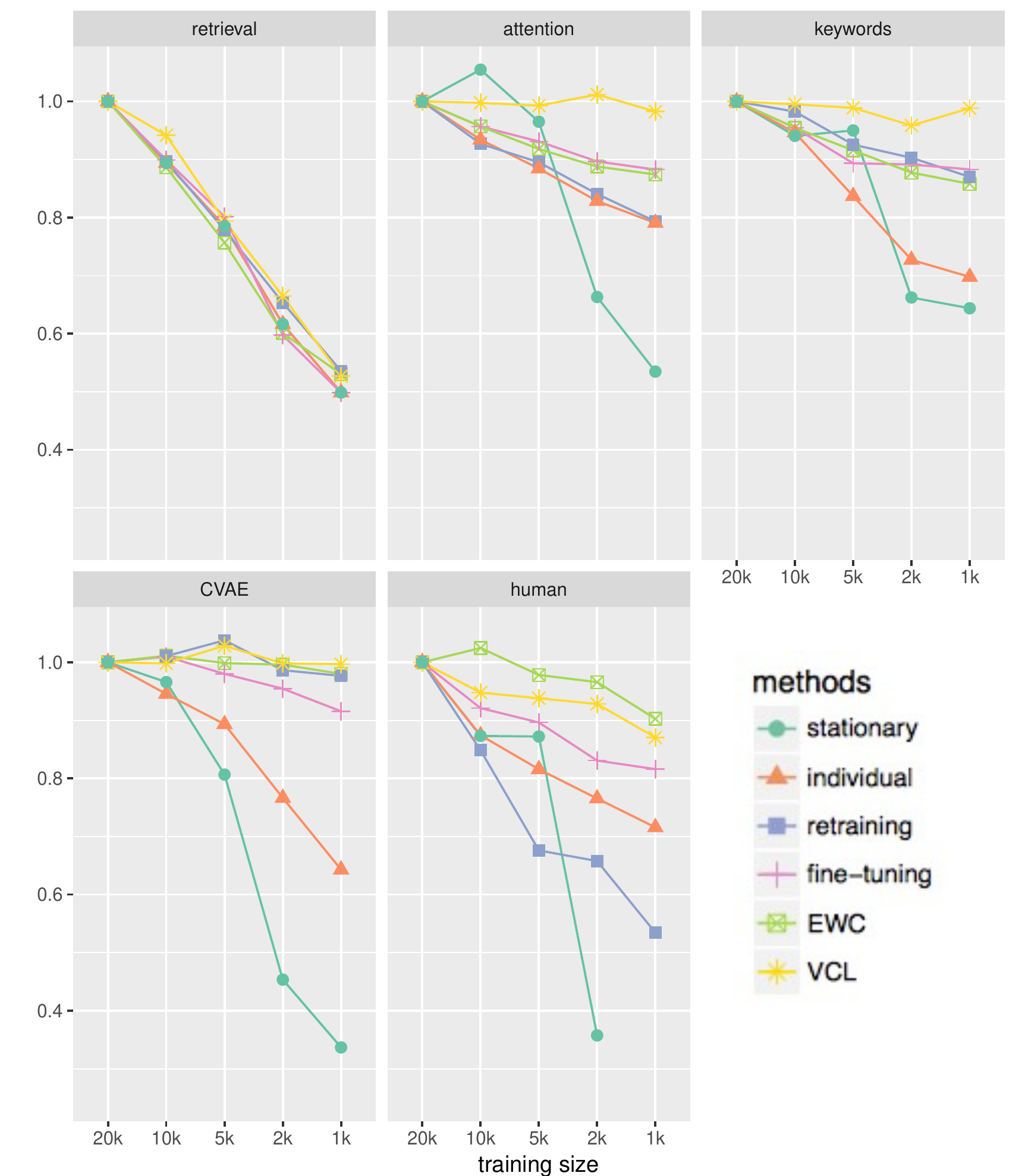}
\caption{\small The normalized change of plasticity when decreasing its training size.}
\label{data_size}
\end{figure}

We investigate the influence on plasticity when decreasing the training size.
As shown in Figure \ref{data_size}, for all learning methods, the plasticity of the first task (retrieval-system) decreases significantly as the training size decreases.
However, with the learning of subsequent tasks, the plasticity of continual learning (EWC and VCL) are less affected by the reduction of training size. It is because that continual learning based dialogue evaluators continually transfer learned knowledge and only need minor updates to adapt to new tasks as the number of tasks increases. 
It can be seen that, although fine-tuning based dialogue evaluator uses previous parameters as an initial value, the evaluation effectiveness of the final task (human-system) still declines seriously with the training size decreases.

\section{Related Work}

\subsection{Automatic Evaluation for Chatbots}

Generally speaking, dialogue systems can be divided into task-oriented ones \cite{walker1997paradise, moller2006memo} and chatbots (a.k.a. open-domain dialogue systems).
In this paper, we focus on the automatic evaluation of open-domain dialogue systems. 
Previous works show that existing unsupervised metrics (like BLEU, ROUGE, and Perplexity) are not applicable for evaluating open-domain dialogue systems \cite{serban2016building, liu2016not}.
Though the retrieval-based dialogue system can be automatically evaluated using precision and recall \cite{zhou2018multi}, that metrics cannot be extended for the generation-based ones.


Therefore, very recent attempts formulate automatic dialogue evaluation as a learning problem.
Inspired by GAN \cite{goodfellow2014generative}, Kannan et al. \shortcite{kannan2017adversarial} train a discriminator to distinguish whether the response is human-generated or not.
Lowe et al., \shortcite{lowe2017towards} propose an end-to-end automatic dialogue evaluation model (ADEM) to calculate human-like scores by triple similarity jointly considering the context, ground-truth, and response.
Tao et al., \shortcite{tao2017ruber} combine a referenced metric, the embedding similarity between ground-truth and response, and an unreferenced metric, the matching score between context and response, by simple heuristics.

\subsection{Continual learning}
Continual learning methods for neural networks are broadly partitioned into three groups of approaches.
\emph{Architectural} approaches \cite{rusu2016progressive, fernando2017pathnet} alter the architecture of the network when every new task emerges. 
It makes architectural complexity grow with the number of tasks.
\emph{Functional} approaches \cite{jung2016less} add a regularization term to the objective, to penalize changes in the input-output function of the neural network. This results in expensive computation as it requires computing a forward pass through the old task's network for every new data point.

The third, \emph{regularization} approaches add constraints to the update of network weights. 
Kirkpatrick et al. \shortcite{kirkpatrick2017overcoming} measure the importance of weights by a point estimate of Fisher information and slow down leaning on weights which are important for old tasks.
Nguyen et al. \shortcite{nguyen2017variational} propose variational continual learning (VCL) in a Bayesian framework. Prior knowledge is represented as a probability density function learning from previous tasks. And the posterior is updated from prior in the light of a new task at the cost of KL loss. 
In this work, we introduce those two model-agnostic regularization approaches to update dialogue evaluators for dialogue systems.

\section{Conclusion}
In this paper, we study the problem of automatic dialogue evaluation from the perspective of continual learning.
Our work is inspired by the observation of the weak generalization of existing neural dialogue evaluators and we propose to alleviate that issue via selectively adapting the evaluator, to jointly fit dialogue systems have been and to be evaluated.
Experimental results show that our continual evaluators are able to adapt to achieve comparable performance with evaluator reconstruction, while our continual evaluators require significantly fewer annotations and eliminate the trouble of maintaining an increasing size of evaluators or annotations.

\bibliography{emnlp-ijcnlp-2019-lilu}
\bibliographystyle{acl_natbib}

\end{document}